\documentclass[11pt,letterpaper]{article}
\usepackage[nohyperref]{emnlp2017}
\usepackage{times}
\usepackage{amsmath}
\usepackage{amssymb}
\usepackage{latexsym}
\usepackage{graphicx}
\usepackage{framed}
\usepackage{multirow}
\usepackage[draft]{hyperref}

\emnlpfinalcopy

\def\emnlppaperid{343}

\DeclareMathOperator*{\argmax}{arg\,max}

\title{Global Normalization of Convolutional Neural Networks \\for 
Joint Entity and Relation Classification}

 \author{Heike Adel \and Hinrich Sch\"{u}tze \\
         Center for Information and Language Processing (CIS)\\
         LMU Munich, Germany\\
         \texttt{heike@cis.lmu.de}}
\date{}

\begin{document}

\maketitle

\begin{abstract}
We introduce globally normalized convolutional
neural networks for 
joint entity classification and relation extraction.
In particular, we propose a way to utilize a linear-chain conditional
random field output layer for predicting
entity types and relations between entities at the same time.
Our experiments show that global normalization outperforms
a locally normalized softmax layer on a benchmark dataset.
\end{abstract}

\section{Introduction}
Named entity classification (EC) and relation extraction (RE)
are important topics in natural language processing.
They are relevant, e.g., for populating knowledge
bases or answering questions from text,
such as ``Where does X live?''

\begin{figure}
\begin{framed}
\footnotesize
\begin{center}
\includegraphics[width=\textwidth]{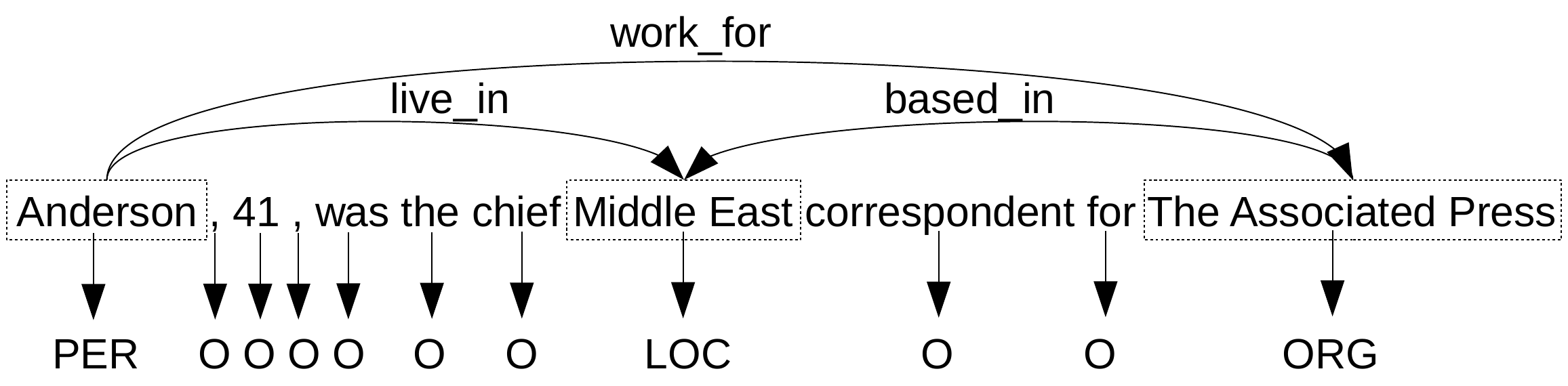}
\end{center}
Model inputs (query entity pairs) $\Rightarrow$ Model outputs:\\
(``Anderson'', ``,'') $\Rightarrow$ PER - N - O\\
(``Anderson'', ``41'') $\Rightarrow$ PER - N - O\\
...\\
(``Anderson'', ``chief'') $\Rightarrow$ PER - N - O\\
(``Anderson'', ``Middle East'') $\Rightarrow$ PER - live\_in - LOC\\
...\\
(``was'', ``for'') $\Rightarrow$ O - N - O\\
...\\
(``for'', ``The Associated Press'') $\Rightarrow$ O - N - ORG\\
\end{framed}
\caption{Examples of our task}
\label{fig:exampleTask}
\end{figure}

Most approaches consider the two tasks
independent from each other or treat them
as a sequential pipeline by first applying a named
entity recognition tool and then classifying
relations between entity pairs.
However, named entity
types and relations are often mutually dependent.
If the types of entities are known, the search space of possible
relations between them can be reduced and vice versa.
This can help, for example, to resolve ambiguities,
such as in the case of ``Mercedes'', which can 
be a person, organization and location.
However, knowing that in the given context, it is
the second argument for the relation
``live\_in'' helps concluding that it is 
a location.
Therefore, we propose a single neural network (NN) for both tasks. 
In contrast to
joint training and multitask learning,
which calculate task-wise costs,
we propose to learn a \emph{joint classification layer}
which is \emph{globally normalized} on the outputs of
both tasks.
In particular, we train the NN parameters based on
the loss of a linear-chain conditional random field (CRF) \cite{crfOrig}.
CRF layers for NNs have been
introduced for token-labeling tasks like
named entity recognition (NER) or part-of-speech tagging \cite{cw2011,crfCode,googleGlobal}.
Instead of labeling each input token as in previous work, 
we model the joint entity and relation classification
problem as a sequence of length three for the CRF layer.
In particular,
we identify the types of two candidate entities (words
or short phrases) given a
sentence 
(we call this entity classification
to distinguish it from the token-labeling
task NER) 
as well as the relation between them.
To the best of our knowledge, this architecture
for combining entity and relation classification
in a single neural network is novel.
Figure \ref{fig:exampleTask} shows
an example of how we model the task:
For each sentence, candidate entities are identified.
Every possible combination of
candidate entities (query entity pair) then forms the input to
our model which predicts the classes for the
two query entities as well as for the relation between them.

To sum up, our contributions are
as follows: 
We introduce
globally normalized convolutional neural
networks for a sentence classification task. In particular,
we present an architecture which allows us to
model joint entity
and relation classification with a single neural network
and classify entities and relations at the same
time, normalizing their scores globally. 
Our experiments confirm that a CNN with a CRF output
layer outperforms a CNN with locally normalized softmax layers.
Our source code is available at \url{http://cistern.cis.lmu.de}.

\section{Related Work}
Some work 
on joint entity and relation classification 
uses distant supervision
for building their own datasets, e.g., \cite{yao,jointEACL}. 
Other studies, which
are described in more detail in the following, use the
``entity and relation recognition'' (ERR) dataset from \cite{data,data2} as we do in this paper.
\newcite{data} develop constraints and use linear programming 
to globally normalize entity types and relations.
\newcite{giuliano} use entity type information
for relation extraction but do not train both
tasks jointly.
\newcite{kate} train task-specific support vector machines and
develop a card-pyramid parsing algorithm to 
jointly model both tasks.
\newcite{miwa2014} use the same dataset but
model the tasks as a table filling problem (see Section \ref{sec:setup2}).
Their model uses both a local and a global
scoring function.
Recently, \newcite{pankaj2016} apply recurrent neural
networks to fill the table. They train them
in a multitask fashion. Previous work also
uses a variety of linguistic features, such
as part-of-speech tags.
In contrast, we use convolutional neural networks
and only word embeddings as input. 
Furthermore, we are the first
to adopt global normalization of neural networks
for this task.

Several studies propose different variants of non-neural CRF models 
for information
extraction tasks but model them as token-labeling
problems \cite{crfSkip,semiCRF,culotta2006,zhu20052d,peng2006information}.
In contrast, we propose a simpler linear-chain
CRF model which directly connects entity and relation classes
instead of assigning a label to each
token of the input sequence.
This is more similar to the factor graph by \newcite{yao}
but computationally simpler.
\newcite{cnnCrf} also apply a CRF layer
on top of continuous representations obtained
by a CNN. However, they use it for
a token labeling task (semantic slot filling)
while we apply the model to a sentence
classification task, motivated by the fact
that a CNN creates single representations
for whole phrases or sentences.

\section{Model}

\begin{figure*}
\centering
\includegraphics[width=.9\textwidth]{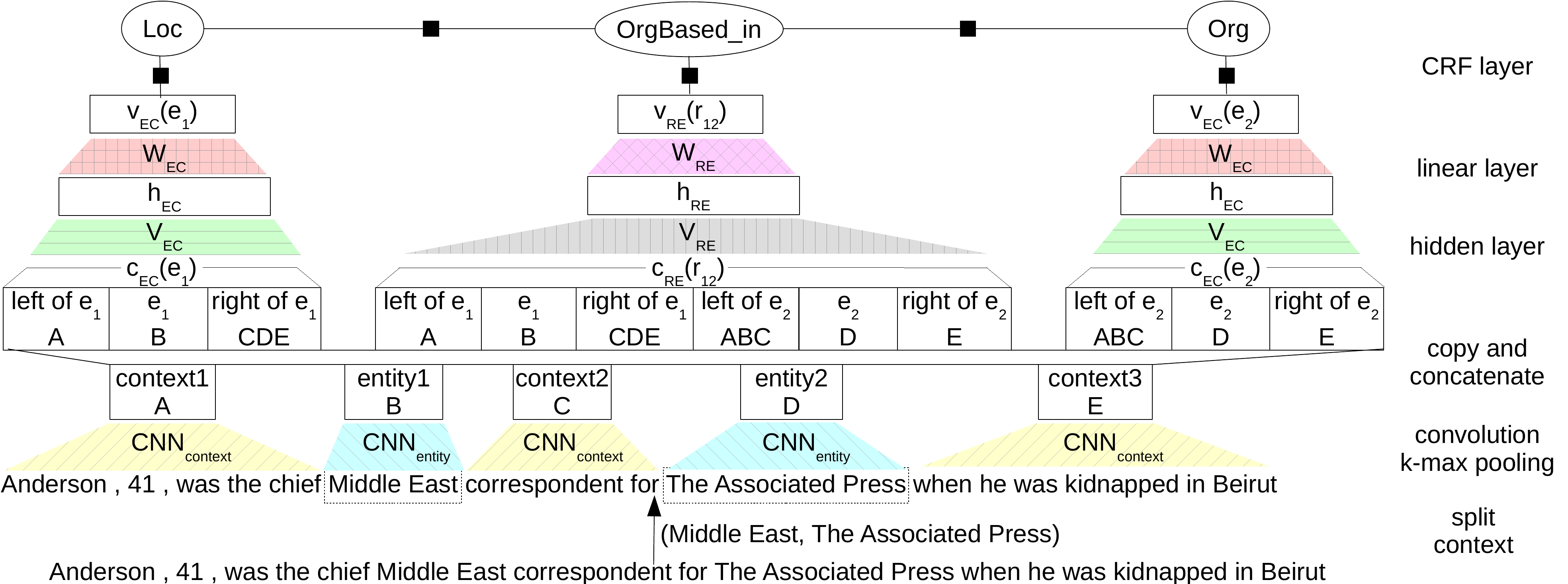}
\caption{Model overview; the colors/shades show which model parts share parameters}
\label{fig:model}
\end{figure*}

\subsection{Modeling Context and Entities}
Figure \ref{fig:model} illustrates our model.

\textbf{Input.}
Given an input sentence and two query entities,
our model
identifies the \emph{types of the entities}
and the \emph{relation} between them; 
see Figure \ref{fig:exampleTask}.
The input tokens are represented
by word embeddings trained on Wikipedia with word2vec \cite{word2vec}.
For identifying the class of an entity $e_k$, the 
model uses the context to its left, the
words constituting $e_k$ and the context to its
right.
For classifying the relation between two
entities $e_i$ and $e_j$, the sentence is split into six parts:
left of $e_i$, $e_i$, right of $e_i$, left of $e_j$, 
$e_j$, right of $e_j$.\footnote{The ERR dataset we use provides boundaries
for entities to concentrate
on the classification task \cite{data}.}
For the example sentence in Figure \ref{fig:exampleTask} and 
the entity pair (``Anderson'', ``chief''),
the context split is: 
$[]$ $[$Anderson$]$ $[$, 41 , was the chief Middle ...$]$
$[$Anderson , 41 , was the$]$ $[$chief$]$ 
$[$Middle East correspondent for ...$]$

\textbf{Sentence Representation.}
For representing the different parts of the input sentence, we use 
convolutional neural networks (CNNs).
CNNs are suitable for RE since a relation
is usually expressed by the semantics
of a whole phrase or sentence.
Moreover, they have proven effective
for RE in previous work \cite{ourNaacl2016}.
We train one CNN layer for 
convolving the entities 
and one for the contexts. 
Using two CNN layers instead of one gives
our model more flexibility.
Since entities
are usually shorter than contexts, the filter width
for entities can be smaller than for
contexts. 
Furthermore,
this architecture simplifies changing the entity representation
from words to characters in future work.

After convolution, we apply $k$-max pooling
for both the entities and the contexts
and concatenate the 
results. The concatenated vector $c_{z}\in\mathbb{R}^{C_{z}}$, 
$z\in\left\{EC, RE\right\}$
is forwarded 
to a task-specific hidden layer 
of size $H_z$
which learns patterns across
the different input parts:
\begin{equation}
h_{z} = \tanh(V_{z}^Tc_{z} + b_{z})
\end{equation}
with 
weights 
$V_{z} \in \mathbb{R}^{C_{z} \times H_{z}}$ and
bias 
$b_{z} \in \mathbb{R}^{H_{z}}$.

\subsection{Global Normalization Layer}
For global normalization, we adopt the
linear-chain
CRF layer by 
\newcite{crfCode}.\footnote{https://github.com/glample/tagger}
It expects scores for
the different classes as input. Therefore, we
apply a linear layer first which
maps the representations $h_{z} \in \mathbb{R}^{H_{z}}$ 
to a vector $v_z$ of the size of the output classes $N = N_{EC} + N_{RE}$:
\begin{equation}
v_{z} = W_{z}^Th_{z}
\end{equation}
with $W_{z} \in \mathbb{R}^{H_{z} \times N}$.
For a sentence classification task,
the input sequence for the CRF layer
is not inherentely clear. 
Therefore, we propose to model the 
joint entity and relation classification problem 
with the following sequence of 
scores (cf., Figure \ref{fig:model}):
\begin{equation}
d = [v_{EC}(e_1),v_{RE}(r_{12}),v_{EC}(e_2)]
\end{equation}
with $r_{ij}$ being
the relation between $e_i$ und $e_j$.
Thus, we approximate the joint probability
of entity types $T_{e_1}$, $T_{e_2}$
and relations $R_{e_1e_2}$ as follows:
\begin{align}
\begin{split}
 & P(T_{e_1}R_{e_1e_2}T_{e_2})  \\
 \approx
 & P(T_{e_1}) \cdot P(R_{e_1e_2}|T_{e_1}) \cdot P(T_{e_2}|R_{e_1e_2})
 \end{split}
\end{align}
Our intuition is that the dependence between relation and
entities is stronger than the dependence between the two entities.

The CRF layer pads its input of length $n = 3$ with begin
and end tags and computes the following score for
a sequence of predictions $y$:
\begin{equation}
s(y) = \sum_{i=0}^{n}Q_{y_iy_{i+1}} + \sum_{i=1}^nd_{i,y_i}
\end{equation}
with $Q_{k,l}$ being the transition score from class $k$ to
class $l$ and 
$d_{p,q}$ being the score of class $q$ at 
position $p$
in the sequence.
The scores are summed because all the 
variables of the CRF layer live in the log space.
The matrix of transition scores $Q \in \mathbb{R}^{(N+2) \times (N+2)}$
is learned during training.\footnote{2 is added
because of the padded begin and end tag}
For training, the forward algorithm 
computes the scores for all possible label sequences $Y$
to get the log-probability of the correct label
sequence $\hat{y}$:
\begin{equation}
log(p(\hat{y})) = \frac{e^{s(\hat{y})}}{\sum_{\tilde{y}\in Y}e^{s(\tilde{y})}}
\end{equation}
For testing, Viterbi is applied to
obtain the label sequence $y^*$ with the maximum score:
\begin{equation}
y^* = \argmax_{\tilde{y} \in Y}s(\tilde{y})
\end{equation}

\section{Experiments and Analysis}
\subsection{Data and Evaluation Measure}
We use the ``entity and relation recognition'' (ERR) dataset from 
\cite{data}\footnote{http://cogcomp.cs.illinois.edu/page/resource\_view/43}
with the train-test split by
\newcite{pankaj2016}.
We tune the parameters on a held-out part
of train.
The data is labeled 
with entity types
and relations (see Table \ref{tab:results}). 
For entity pairs without a relation,
we use the label N.
Dataset statistics 
and model parameters are provided in the appendix.

Following previous work, we compute $F_1$
of the individual classes
for EC and RE, 
as well as a task-wise macro $F_1$ score. 
We also report the average of scores across tasks (Avg EC+RE).

\subsection{Experimental Setups}
\textbf{Setup 1: Entity Pair Relations.}
\newcite{data,data2,kate}
train separate models for EC and 
RE on the ERR dataset.
For RE, they only identify relations
between named entity pairs.
In this setup, the query entities for our
model are only named entity pairs.
Note that this facilitates EC
in our experiments.

\begin{figure}
\centering
\includegraphics[width=.4\textwidth]{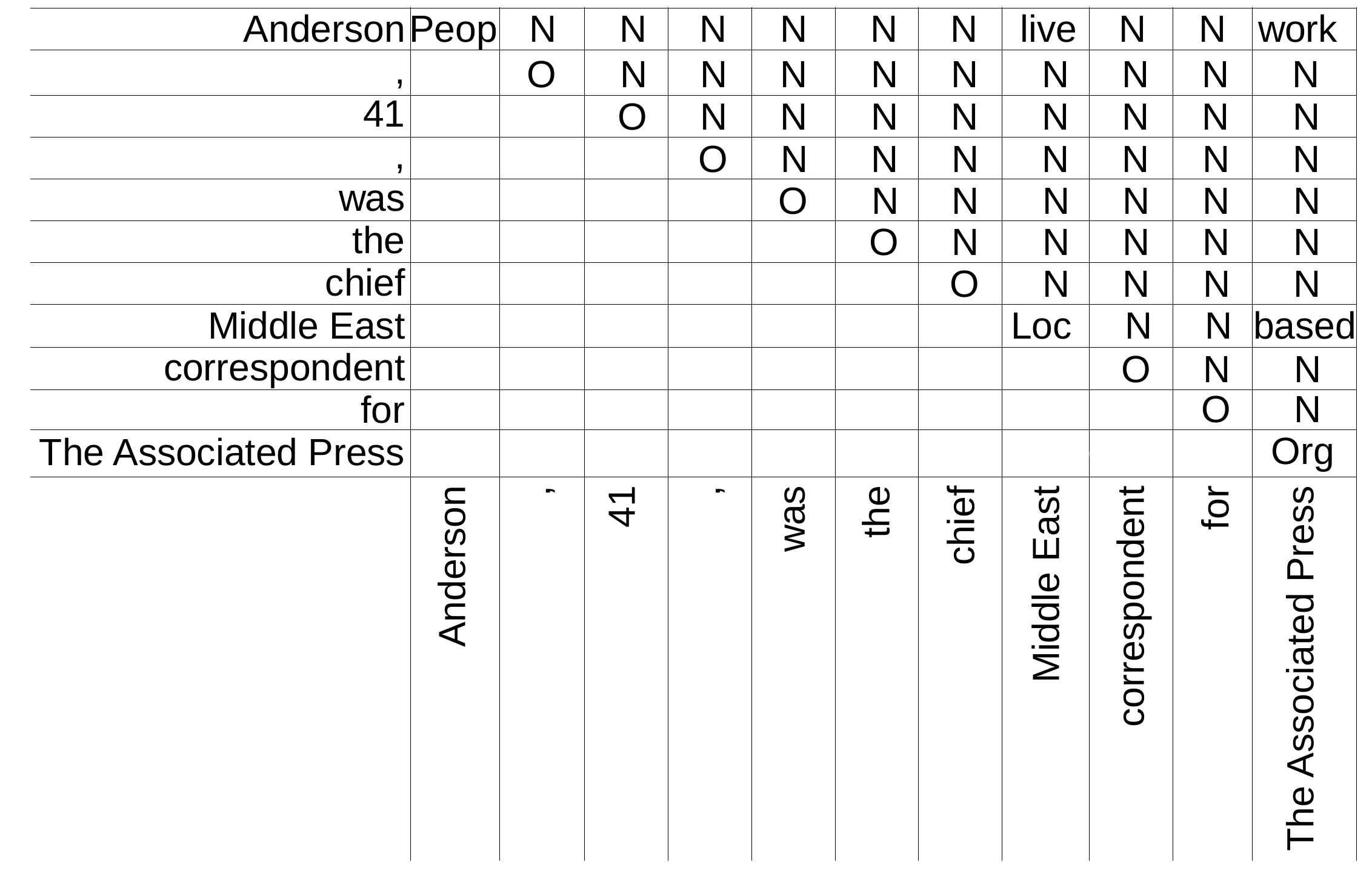}
\caption{Entity-relation table} 
\label{fig:tablefilling}
\end{figure}

\textbf{Setup 2: Table Filling.}
\label{sec:setup2}
Following \newcite{miwa2014,pankaj2016}, we also model 
the joint task of EC and RE
as a table filling task.
For a sentence with length $m$, we
create a quadratic table. Cell $(i,j)$
contains the relation between
word $i$ and word $j$ (or N for no relation).
A diagonal cell $(k,k)$ contains the
entity type of word $k$.
Following previous work, we only predict classes
for half of the table, i.e. for $m(m+1)/2$ cells.
Figure \ref{fig:tablefilling} shows
the table for the example sentence from Figure \ref{fig:exampleTask}.
In this setup,
each cell $(i,j)$ with $i \ne j$ is a separate input query to our model.
Our model outputs a prediction for cell $(i,j)$ (the 
relation between $i$ and $j$) and predictions for cells $(i,i)$ and 
$(j,j)$ (the types of $i$ and $j$).
To fill the diagonal with entity classes, we aggregate
all predictions for the particular entity by using majority vote.
Section \ref{sec:analysis1} shows that the individual predictions
agree with the majority vote in almost all cases.

\textbf{Setup 3: Table Filling Without Entity Boundaries.}
The table from setup 2 includes one row/column per multi-token entity,
utilizing the given entity boundaries of the ERR dataset.
In order to investigate the impact of the entity boundaries on the
classification results, we also consider another table filling setup
where we ignore the boundaries and assign one row/column per token.
Note that this setup is also used by prior work on table filling
\cite{miwa2014,pankaj2016}.
For evaluation, we follow \newcite{pankaj2016}
and score a multi-token entity as correct if at least one of its comprising cells
has been classified correctly.

\textbf{Comparison.}
The most important difference between setup 1 and setup 2 is
the number of entity pairs with no relation (test set: $\approx$3k for 
setup 1, $\approx$121k for setup 2). This makes setup 2 more challenging.
The same holds for setup 3 which considers the same number of entity pairs
with no relation as setup 2.
To cope with this, we randomly subsample negative instances
in the train set of setup 2 and 3.
Setup 3 considers the most query entity pairs in total since multi-token entities are
split into their comprising tokens. However, setup 3 represents a more realistic scenario
than setup 1 or setup 2 because in most cases, entity boundaries are not given.
In order to apply setup 1 or 2 to another dataset without entity boundaries,
a preprocessing step, such as entity boundary recognition or chunking
would be required.

\begin{table*}
\centering
\footnotesize
\begin{tabular}{l|c|c||c|c||c|c}
& \multicolumn{2}{c||}{Setup 1} & \multicolumn{2}{c||}{Setup 2} & \multicolumn{2}{c}{Setup 3}\\
& softmax & CRF & softmax & CRF & softmax & CRF\\
\hline
Peop & \textbf{95.24} & 94.95 &  93.99 & \textbf{94.47} & 91.46 & \textbf{92.21}\\
Org & \textbf{88.94} & 87.56 & 78.95 & \textbf{79.37} & 67.29 & \textbf{67.91}\\
Loc & 93.25 & \textbf{93.63} & 90.69 & \textbf{90.80} & 85.99 & \textbf{86.20}\\
Other & \textbf{90.38} & 89.54 & 73.78 & \textbf{73.97} & \textbf{62.67} & 61.19\\
\hline
Avg EC & \textbf{91.95} & 91.42 & 84.35 & \textbf{84.65} & 76.85 & \textbf{76.88}\\
\hline
Located\_in & 55.03 & \textbf{57.72} & 51.03 & \textbf{55.13} & 44.96 & \textbf{52.29}\\
Work\_for & \textbf{71.23} & 70.67 & 52.89 & \textbf{61.42} & 52.63 & \textbf{65.31}\\
OrgBased\_in & 53.25 & \textbf{59.38} & 56.96 & \textbf{59.12} & 46.15 & \textbf{57.65}\\
Live\_in & \textbf{59.57} & 58.94 & \textbf{64.29} & 60.12 & \textbf{64.09} & 61.45\\
Kill & 74.70 & \textbf{79.55} & 69.14 & \textbf{74.73} & \textbf{82.93} & 75.86\\
\hline
Avg RE & 62.76 & \textbf{65.25} & 58.86 & \textbf{62.10} & 58.15 & \textbf{62.51}\\
\hline
Avg EC+RE & 77.36 & \textbf{78.33} & 71.61 & \textbf{73.38} & 67.50 & \textbf{69.69}
\end{tabular}
\caption{$F_1$ results for 
entity classification (EC) and relation extraction (RE)
in the three setups}
\label{tab:results}
\end{table*}

\subsection{Experimental Results}
Table \ref{tab:results} shows the
results of our globally normalized model in comparison to
the same model with locally normalized softmax output layers 
(one for EC and one for RE).
For setup 1, the CRF layer performs comparable
or better than the softmax layer. For
setup 2 and 3, the improvements are more apparent.
We assume that the model can benefit more
from global normalization in the case of table filling because
it is the more challenging setup.
The comparison between setup 2 and setup 3 shows
that the entity classification suffers from
not given entity boundaries (in setup 3). 
A reason could be that the model cannot
convolve the token embeddings of the multi-token
entities anymore when computing the
entity representation (context B and D in Figure \ref{fig:model}).
Nevertheless, the
relation classification performance 
is comparable in setup 2 and setup 3. This shows that the model can
internally account for potentially wrong entity classification results due
to missing entity boundaries.

The overall results (Avg EC+RE) of
the CRF
are better than the results of the softmax 
layer for all three setups.
To sum up, the
improvements of the
linear-chain
CRF 
show that
(i) joint EC and RE benefits
from global normalization and 
(ii) our way of creating the input
sequence for the CRF for joint 
EC and RE is effective.

\textbf{Comparison to State of the Art.}
Table \ref{tab:stateoftheart} shows our results
in the context of
state-of-the-art results:
\cite{data2}, \cite{kate}, 
\cite{miwa2014},
\cite{pankaj2016}.\footnote{We only show results of 
single models, no ensembles. Following previous studies, we omit the
entity class ``Other'' when computing
the EC score.}
Note that the results are not comparable
because of the different setups 
and different train-test splits.\footnote{Our results 
on EC in setup 1 are also not comparable since we only input
named entities into our model.}

Our results are best comparable with \cite{pankaj2016}
since we
use the same setup and train-test splits. However, their
model is more complicated with 
a lot of hand-crafted features and various iterations
of modeling dependencies among entity and
relation classes.
In contrast, we only use pre-trained word embeddings
and train our model end-to-end with
only one iteration per entity pair.
When we compare with their model without
additional features (G et al. 2016 (2)), 
our model performs worse for EC but better
for RE and comparable for Avg EC+RE.

\begin{table}
\footnotesize
\centering
 \begin{tabular}{l|c|c|c|c|c}
 Model & S & Feats & EC & RE & EC+RE\\
 \hline
 R \& Y 2007 & 1 & yes &  85.8 & 58.1 & 72.0\\
 K \& M 2010 & 1 & yes & 91.7 & 62.2 & 77.0\\
 \hline
 Ours (NN CRF) & 1 & no & \textbf{92.1} & \textbf{65.3} & \textbf{78.7}\\
 \hline
 \hline
Ours (NN CRF) & 2 & no & \textbf{88.2} & \textbf{62.1} & \textbf{75.2}\\
\hline
\hline
 M \& S 2014 & 3 & yes & 92.3 & \textbf{71.0} & \textbf{81.7}\\
 G et al. 2016 (1) & 3 & yes & \textbf{92.4} & 69.9 & 81.2\\
\hline
 G et al. 2016 (2) & 3 & no & \textbf{88.8} & 58.3 & \textbf{73.6}\\
 Ours (NN CRF) & 3 & no & 82.1 & \textbf{62.5} & 72.3
 \end{tabular}
\caption{Comparison to state of the art (S: setup)}
\label{tab:stateoftheart}
\end{table}

\subsection{Analysis of Entity Type Aggregation}
\label{sec:analysis1}
As described in Section \ref{sec:setup2},
we aggregate the EC results
by majority vote.
Now, we analyze their disagreement. 
For our best model,
there are only 9 entities (0.12\%) 
with disagreement in the test data. For those, the
max, min and median disagreement with the majority label 
is 
36\%, 2\%, and 8\%,
resp. Thus, the disagreement is 
negligibly small.

\begin{figure}
 \centering
 \includegraphics[width=.4\textwidth]{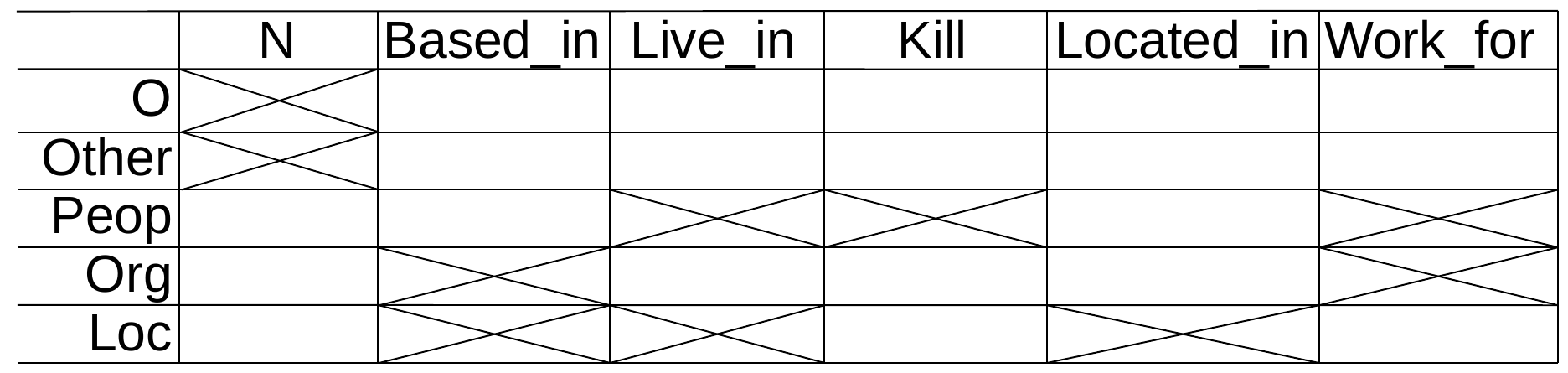}
 \caption{Most strongly correlated entity types and relations
according to CRF transition matrix}
 \label{fig:analysis1}
\end{figure}

\subsection{Analysis of CRF Transition Matrix}
To analyze the CRF layer, we extract which 
transitions have scores above 0.5.
Figure \ref{fig:analysis1} shows that the layer
has learned correct correlations between 
entity types and relations.

\section{Conclusion and Future Work}
In this paper, we presented the first study on global
normalization of
neural networks for a sentence classification task without
transforming it into a token-labeling problem.
We trained a convolutional neural network with a linear-chain
conditional random field output layer on joint entity 
and relation classification 
and showed that it
outperformed using a locally normalized softmax layer.

An interesting future direction is the
extension of the linear-chain CRF to jointly normalize all predictions
for table filling in a single model pass.
Furthermore, we plan to verify our results
on other datasets in future work.

\section*{Acknowledgments}
Heike Adel is a recipient of the Google European
Doctoral Fellowship in Natural Language Processing 
and this research is supported by this fellowship. This work
was also supported by DFG (SCHU 2246/4-2).

\bibliography{emnlp2017}
\bibliographystyle{emnlp_natbib}

\begin{appendix}
\section{Dataset Statistics}
Table \ref{tab:statistics} provides statistics
of the data composition in our different setups which
are described in the paper. The N class of setup 2 and setup 3
has been subsampled in the training and development set 
as described in the paper.

\begin{table}[h]
\centering
\footnotesize
\begin{tabular}{l|r|r|r}
& train & dev & test\\
\hline
Peop & 1146 & 224 & 321\\
Org & 596 & 189 & 198\\
Loc & 1204 & 335 & 427\\
Other & 427 & 110 & 125\\
O &  20338 & 5261 & 6313\\
\hline
Located\_in & 243 & 66 & 94\\
Work\_for & 243 & 82 & 76\\
OrgBased\_in & 239 & 106 & 105\\
Live\_in & 342 & 79 & 100\\
Kill & 203 & 18 & 47\\
N (setup 1) & 10742 & 2614 & 3344\\
N (setup 2/3) & 123453 & 30757 & 120716
\end{tabular}
\caption{Dataset statistics for our different experimental setups}
\label{tab:statistics}
\end{table}

Note that the sum of
numbers of relation labels is slightly different to the numbers
reported in (Roth and Yih, 2004). According to their 
website \url{https://cogcomp.cs.illinois.edu/page/resource_view/43}, 
they have updated the corpus.

\section{Hyperparameters}

\begin{table}[h]
\centering
\footnotesize
 \begin{tabular}{l|l|c|c|c|c}
  Setup & Output layer & $nk_C$ & $nk_E$ & $h_C$ & $h_E$ \\
  \hline
  1 & softmax & 500 & 100 & 100 & 50\\ 
  2 & softmax & 500 & 100 & 100 & 50\\
  3 & softmax & 500 & 100 & 100 & 50\\
\hline
  1 & CRF & 200 & 50 & 100 & 50\\
  2 & CRF & 500 & 100 & 200 & 50\\
  3 & CRF & 500 & 100 & 100 & 50
 \end{tabular}
 \caption{Hyperparameter optimization results}
 \label{tab:params}
\end{table}

Table \ref{tab:params} provides the hyperparameters
we optimized on dev ($nk_C$: number of convolutional filters
for the CNN convolving the contexts, $nk_E$: number of convolutional
filters for the CNN convolving the entities; $h_C$: number of
hidden units for creating the final context representation,
$h_E$: number of hidden units for creating the final 
entity representation). 

For all models, we use a filter width of 3 for the context
CNN and a filter width of 2 for the entity CNN (tuned in prior
experiments and fixed for the optimization
of the parameters in Table \ref{tab:params}).

For training, we apply gradient descent with a batch size
of 10 and an initial learning rate of 0.1. When the performance
on dev decreases, we halve the learning rate. The model is
trained with early stopping on dev, with a maximum number of 20 epochs.
We apply L2 regularization with $\lambda = 10^{-3}$.
\end{appendix}

\end{document}